\documentclass[lettersize,journal]{IEEEtran}
\usepackage{amsmath,amsfonts,amssymb}
\usepackage{algorithmic}
\usepackage{algorithm}
\usepackage{array}
\usepackage[caption=false,font=normalsize,labelfont=sf,textfont=sf]{subfig}
\usepackage{textcomp}
\usepackage{stfloats}
\usepackage{placeins}

\usepackage{url}
\usepackage{verbatim}
\usepackage{graphicx}
\usepackage{cite}
\usepackage{xcolor}
\begin{document}
\bstctlcite{IEEEexample:BSTcontrol}

\title{ZARA: Zero-shot Animatable and Relightable 3D Avatar reconstruction from a single image}

\author{Yixiao Chen, Dongdong Weng, Chengfeng Zhao, Tao Gong, Zhihe Zhao, Zhe Zhu, Yue Liu, and Yuan Liu%
\thanks{Corresponding author: Dongdong Weng, and Yuan Liu.}%
\thanks{Chengfeng Zhao is the project leader for this work.}
}

\markboth{Journal of \LaTeX\ Class Files,~Vol.~14, No.~8, August~2021}{Shell \MakeLowercase{\textit{et al.}}: A Sample Article Using IEEEtran.cls for IEEE Journals}

\maketitle

\begin{abstract}
Recent large human reconstruction models create remarkable animatable 3D avatars from a single image, but their appearance remains entangled with the illumination and cannot be relighted. In contrast, existing 3D relightable-avatar reconstruction methods typically require person-specific optimization. We present ZARA, a zero-shot feed-forward framework that reconstructs an animatable 3D Gaussian avatar together with its intrinsic attributes from a single image. ZARA couples a pretrained large-scale human reconstruction model with a Material Transformer to inherit geometry priors while learning material-aware Gaussian representations. To preserve localized details, we introduce Gaussian-aligned material feature fusion strategy, which injects multi-scale material features onto individual Gaussians and combines them with geometry tokens. As most human-avatar datasets lack ground-truth material maps, we use pseudo-material labels as conditioning inputs and supervision targets during training. We further introduce MatFace regularization to preserve facial structure and color in albedo prediction under unreliable facial pseudo-material supervision. Extensive experiments demonstrate that ZARA reconstructs animatable and relightable humans within tens of seconds without person-specific optimization or post-processing, outperforming existing baselines in both relighting quality and materials generalization ability. As the ability of recovering the physically-based-rendering attributes of the human, ZARA also enables direct material editing of the avatar's appearance.
\end{abstract}


\section{Introduction}

\IEEEPARstart{R}{econstructing} high-fidelity and animatable 3D human avatars with relighting capability is crucial for many applications, such as games, immersive AR/VR, and film production. Most existing methods for reconstructing animatable and relightable avatars are person-specific, that is, a trained model can only be applied to a particular individual wearing a specific outfit. Although these methods have achieved promising rendering quality with the significant development of NeRF~\cite{NeRF,NeuralBody,HumanNeRF} and 3DGS~\cite{3DGS,GaussianAvatar,GauHuman} techniques, they require time-consuming reconstruction or optimization for each subject and require at least monocular video as input, which limits their broader applicability. A series of methods rely on ultra-high-quality OLAT (one-light-at-a-time) data captured using Light Stage systems~\cite{DebevecLightStage,DeepAppearanceModels,RelightableGaussianCodecAvatars}, and employ generative models to learn accurate avatar appearance, material, and motion models. However, they still model specific subject, while the costs of data acquisition and storage are prohibitively high.

Recently, numerous Transformer-based methods have demonstrated the ability to reconstruct relightable 3D objects and scenes from sparse views or even a single image. In human reconstruction, S3F \cite{S3F} has explored reconstructing relightable and animatable implicit 3D avatars by structured 3D features from a single image. On the other hand, large reconstruction model (LRM) based methods\cite{LHM,IDOL,DynaAvatar} which are trained on large-scale datasets enable the reconstruction of animatable 3D avatars from a single image in seconds, and are demonstrated remarkable zero-shot 3D human reconstruction quality. LCA~\cite{LCA} further demonstrates that a pretrained feed-forward large Gaussian avatar model can be extended to relighting through dedicated post-training using OLAT dataset.

Although encouraging results are achieved, however, S3F\cite{S3F} leverages an albedo-shading-based relighting which lacks explicit physically based rendering (PBR) material decomposition, limiting flexibly modeling of material-illumination interaction. For non-relightable LRM-based methods \cite{LHM,IDOL,DynaAvatar}, one possible solution is to first reconstruct an animatable avatar and then apply a video relighting model, however, this two-stage pipeline introduces an additional inference stage and does not recover explicit 3D PBR materials. In contrast, LCA\cite{LCA} requires controlled multi-view, time-multiplexed light-stage data for its relighting-specific post-training. Besides, existing human-avatar datasets predominantly provide RGB images, some even capture the same subject and pose under different lighting conditions. However, large-scale ground-truth material annotations remain scarce, preventing direct supervision with material maps.

In this work, we propose ZARA, a zero-shot feed-forward framework that reconstructs an animatable 3D Gaussian avatar together with explicit per-Gaussian PBR attributes from a single image, enabling direct relighting. ZARA builds on a pretrained large human reconstruction model (LHRM), which generates human shape and is efficiently adapted using LoRA~\cite{LoRA}, while a Material Transformer trained from scratch learns intrinsic attributes. A ConvNext style material encoder~\cite{ConvNeXt} extracts multi-scale features from the reference material maps constructing global material tokens. Geometry tokens from LHRM serve as spatial queries, and global material tokens provide long-range material context, allowing the model to predict intrinsic properties without person-specific optimization or post-processing.

To retain localized material evidence, we introduce Gaussian-aligned material feature fusion strategy. The canonical Gaussian avatars are deformed to the reference pose and projected into the reference camera, after which multi-scale material features and raw Material maps are assigned to individual Gaussians. These local features are fused and injected into the corresponding per-gaussian geometry tokens through learned gates. The resulting material attributes combine global material semantics with explicit pixel-to-Gaussian evidence, which is important for preserving high-frequency texture and material boundaries.

A straightforward strategy for material learning relies on the inverse-rendering framework: the model predicts the PBR attributes, and yields RGB outputs through the rendering equation for self-supervision. Nevertheless, this approach is plagued by the ill-posed nature of inverse-rendering, which impedes training. We therefore resort to direct material supervision. Given most existing human-avatar datasets lack ground-truth material maps, inspired by MvInverse\cite{MVInverse} and UniRelight \cite{UniRelight}, we construct pseudo-material maps through a frozen inverse-rendering model~\cite{DiffusionRenderer}, with the resulting maps serving as conditioning priors and supervision targets. Furthermore, because the facial regions of pseudo-label may be misaligned, we introduce MatFace Regularization for facial albedo supervision. Instead of relying exclusively on facial pseudo-albedo-label pixels that may contain identity shifts, MatFace Regularization uses the RGB image as structural guidance and the pseudo-label albedo as the color source, which preserves recognizable facial features while learning intrinsic facial materials.

Together, these components enable ZARA to reconstruct relightable and animatable avatars impressively in tens of seconds from a single image. Extensive experiments demonstrate that our framework improves the generalization and realism of zero-shot 3D animatable and relightable avatar generation, bridging the gap between reconstruction under fixed lighting conditions and relighting. Since our method explicitly generates PBR materials for 3D Gaussian avatars, it also supports appearance editing via material modification.

\begin{table}[!t]
\caption{Comparison of existing avatar reconstruction methods and the proposed ZARA. Each column indicates whether the method reconstructs avatars in a zero-shot manner, whether it supports relighting, whether it generates PBR materials, and whether it can be relighted form single image input.}
\label{tab:method-comparison}
\centering
\footnotesize
\setlength{\tabcolsep}{4pt}
\renewcommand{\arraystretch}{1.22}
\setlength{\extrarowheight}{1.5pt}
\newcommand{\tabyes}{\textcolor{green!65!black}{\ensuremath{\checkmark}}}
\newcommand{\tabno}{\textcolor{red!85!black}{\ensuremath{\times}}}
\begin{tabular}{lcccc}
\hline
\noalign{\vskip 2pt}
Method & \shortstack{Zero-\\Shot} & Relightable & PBR & \shortstack{Single-image\\Relighting} \\
\noalign{\vskip 2pt}
\hline
DNF\cite{DNFAvatar}        & \tabno  & \tabyes & \tabyes & \tabno  \\
SG-IA\cite{SGIA}       & \tabno  & \tabyes & \tabyes & \tabno  \\
IDOL\cite{IDOL}        & \tabyes & \tabno  & \tabno  & --  \\
LHM\cite{LHM}         & \tabyes & \tabno  & \tabno  & --  \\
S3F\cite{S3F}         & \tabyes & \tabyes & \tabno  & \tabyes  \\
LCA\cite{LCA}         & \tabyes & \tabyes & \tabyes & N/R  \\
\textbf{Ours}        & \tabyes & \tabyes & \tabyes & \tabyes \\
\hline
\end{tabular}
\end{table}

Our main contributions are summarized as follows:

\begin{itemize}
    \item We propose ZARA, a zero-shot feed-forward framework that reconstructs an animatable and relightable 3D Gaussian avatar with explicit per-Gaussian PBR attributes from a single image.
    \item We introduce a cascaded geometry-material architecture that couples a pretrained Geometry Transformer with a Material Transformer, thereby integrating geometric tokens, global material context, and per-gaussian aligned local material features for each Gaussian.
    \item We introduce MatFace Regularization, which combines structural cues from the input RGB image with color cues from pseudo-label albedo to preserve facial structure and intrinsic color under unreliable pseudo-label material supervision.
\end{itemize}

\section{Related Work}

\subsection{PBR-Based Relighting and Inverse Rendering}

Relighting aims to synthesize the appearance of a subject under a target illumination. Image-based relighting methods\cite{ICLight,UniRelight} directly transform the observed appearance according to the target illumination, without necessarily recovering explicit 3D geometry or material attributes. However, nearly all these methods perform relighting in image space and therefore struggle to maintain consistency across viewpoints. On the other hand, inverse rendering\cite{PhySG,NeRD,InvRenderIndoor,NeILF,TensoIR,DiffusionRenderer,MVInverse} addresses the inverse problem of decomposing observed images into geometry, materials, and illumination, thereby providing the intrinsic properties required by the physically based rendering. Recent 3D relightable generation methods\cite{GSIR,Relightable3DGaussians,RelitLRM,LIRM,LMGM,Hunyuan3D21} further recover or assign physically based material attributes to objects or scenes, enabling them to be rendered under novel environment illumination with the multi-view consistency. However, these methods mainly focus on static objects or scenes and do not explicitly address the reconstruction of animatable and relightable 3D full-body avatars from a single image.
\begin{figure*}[!t]
    \centering
    \includegraphics[width=\textwidth]{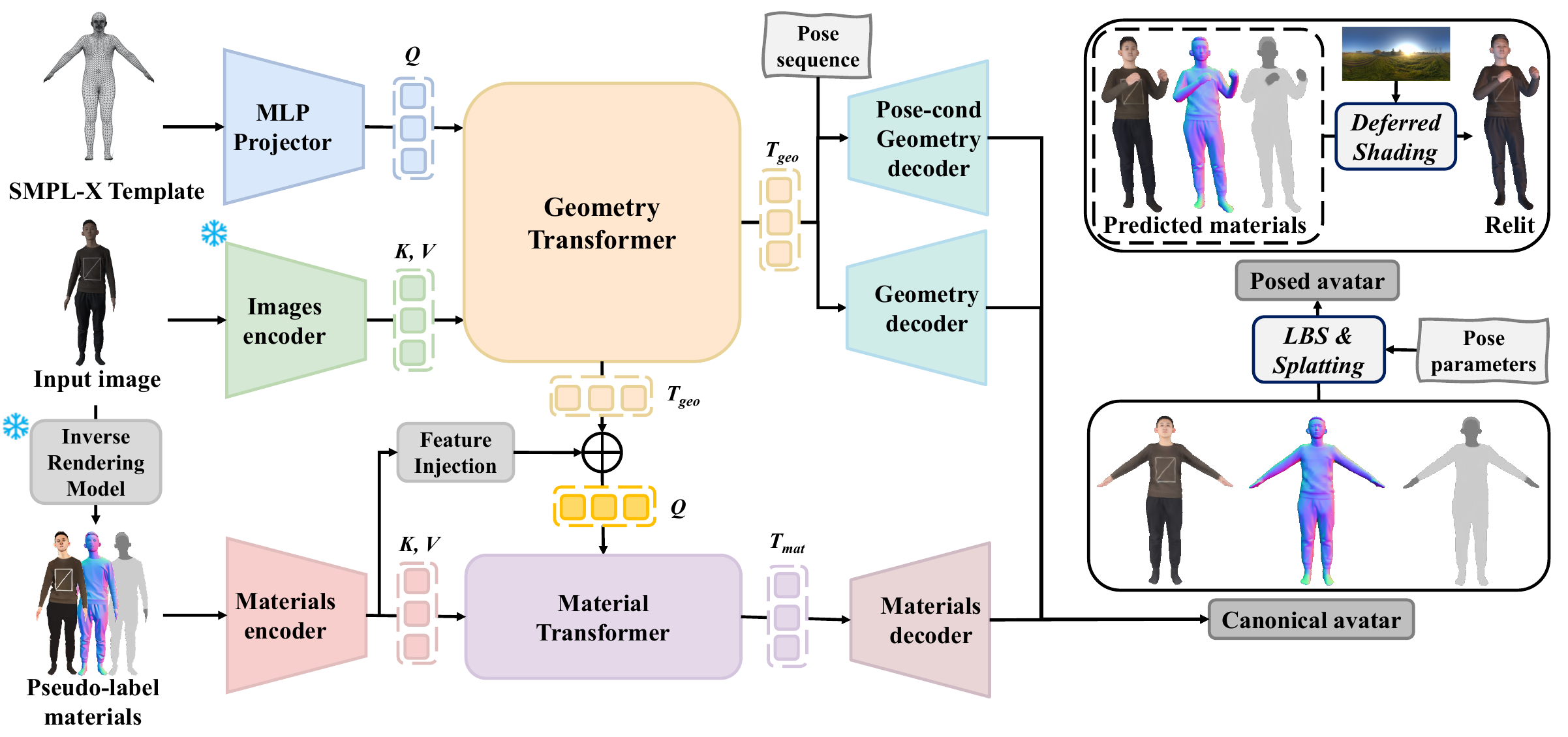}
    \caption{Overview of ZARA and its core network components. Given a single image, the pretrained geometry branch reconstructs canonical Gaussian geometry, while the material branch encodes material maps and predicts per-Gaussian intrinsic attributes. The Gaussian decoding branches predict geometry, pose-dependent geometry offsets, and materials. The material encoder receives the prepared material priors and extracts features that are fused into global material tokens and per-Gaussian material features. The resulting canonical avatar is animated by the target motion and rendered under novel illumination.}
    \label{fig:pipeline-overview}
    \label{fig:gaussian-decoder}
\end{figure*}

\subsection{Relightable and Animatable Avatar Reconstruction}

Relightable avatar methods~\cite{Relighting4D,RelightableNeuralAvatarVideos,NeuralRelightableHuman,RelightableAvatarSparseView} model geometry, materials, and illumination to animate avatars and render them under novel lighting conditions. RANA\cite{RANA} reconstructs relightable articulated neural avatars from monocular videos by estimating geometry, albedo, and lighting. ARG\cite{ARG} extends Gaussian-based avatar modeling with PBR attributes using a Style-UNet architecture. SG-IA\cite{SGIA} introduces surfel-based Gaussian inverse rendering to estimate PBR properties for dynamic clothed humans from monocular videos. DNF-Avatar\cite{DNFAvatar} distills neural fields into an explicit Gaussian representation for real-time animatable avatar relighting. More recent methods\cite{RelightableGaussianCodecAvatarsHead,URAvatar,
RelightableGaussianCodecAvatars,RelightableNeuralActor,
GRA,DRex,RHC,EgoRelight} significantly improve relighting realism by combining generative-model based controllable avatar representations with intrinsic or albedo-like appearance cues and learned PBR-like or illumination-conditioned rendering from controlled captures. However, all of above approaches are person-specific and require monocular or multi-view videos, controlled capture setups, OLAT data, or costly per-subject optimization. 

\subsection{Feed-Forward single image Animatable Avatar Reconstruction}

Recent feed-forward LRM-based methods \cite{LHM,IDOL,AniGS,PERSONA} have substantially improved the quality and reconstruction speed of animatable avatars from a single image. Instead of optimizing a separate representation for each subject, these methods learn strong human priors from large-scale datasets and directly predict animatable 3D avatars at inference time. LHM\cite{LHM} reconstructs animatable 3D Gaussian avatars within seconds using MM-Transformer-based large human reconstruction models. IDOL\cite{IDOL} performs avatar creation from a single image in UV space by transformer-based framework with large generated multi-view human data. DynaAvatar\cite{DynaAvatar} further expands LHM models with motion-dependent cloth dynamics by cascading a motion-window processing module in a zero-shot framework. Other methods\cite{AniGS,PERSONA} use generative multi-view priors or diffusion-generated pose-rich supervision to improve geometric detail, identity preservation, and pose-driven animation quality. Despite their advances, above feed-forward 3D gaussian avatar methods focus on reconstructing geometry, deformation, and appearance under the observed illumination. The recovered appearance of these methods therefore often remains entangled with the lighting condition of the input image, and these methods do not explicitly estimate intrinsic properties or equip with relighting module. As a result, the reconstructed avatars cannot be directly relighted under novel illumination.

Among the few feed-forward alternatives with relighting capability, LCA~\cite{LCA} extends its LHRM-based pretrained avatar model to relighting by replacing its decoder heads and performing relighting-specific post-training on controlled multi-view captures with OLAT data using learnable radiance transfer. S3F~\cite{S3F} reconstructs animatable implicit avatars with structured body-aligned 3D features and performs relighting with albedo and shading, which represents material-light interactions jointly within illumination code rather than through independently parameterized reflectance attributes. Changing the lighting requires re-running the network to obtain the code, which limiting the ability of material-illumination interaction. ZARA instead estimate pbr materials to canonical Gaussians from a single image and enable relighted appearance through deferred PBR for each target pose and view, providing explicit material control and physically grounded, view-consistent relighting without person-specific optimization or post-training.

\section{Preliminaries}

\subsection{Physically Based Rendering}
\label{sec:pbr-rendering}

Physically based rendering computes the outgoing radiance at a surface point $\mathbf{x}$ by integrating the incident illumination over the visible hemisphere~\cite{Kajiya1986RenderingEquation}. Following the microfacet formulation used in \cite{CookTorrance1982,Walter2007Microfacet,Burley2012Disney}, the outgoing radiance toward direction $\boldsymbol{\omega}_o$ is
\begin{equation}
    L_o(\mathbf{x},\boldsymbol{\omega}_o)
    = \int_{\Omega^+}
    L_i(\mathbf{x},\boldsymbol{\omega}_i)
    f_r(\boldsymbol{\omega}_i,\boldsymbol{\omega}_o;\mathbf{x})
    [\mathbf{n}^{\mathsf T}\boldsymbol{\omega}_i]_+
    \,\mathrm{d}\boldsymbol{\omega}_i,
    \label{eq:rendering-equation}
\end{equation}
where $L_i$ is the incident radiance which is represented by environment map, $\mathbf{n}$ is the unit surface normal, and $[z]_+=\max(z,0)$. We use a simplified Disney microfacet BRDF \cite{Burley2012Disney} consisting of Lambertian diffuse and GGX specular terms,
\begin{equation}
\begin{split}
    f_r &= f_d + f_s, \qquad f_d=\frac{\mathbf{a}}{\pi},\\
    f_s &=
    \frac{D_{\mathrm{GGX}}(\mathbf{h};r)\,
          F(\boldsymbol{\omega}_o,\mathbf{h};F_0)\,
          G(\boldsymbol{\omega}_i,\boldsymbol{\omega}_o;r)}
         {4[\mathbf{n}^{\mathsf T}\boldsymbol{\omega}_i]_+
            [\mathbf{n}^{\mathsf T}\boldsymbol{\omega}_o]_+},
\end{split}
\label{eq:microfacet-brdf}
\end{equation}
where \(\mathbf{a}\), \(\mathbf{r}\), and \(\mathbf{n}\) denote albedo, roughness, and normal, respectively. $\mathbf{h}$ is the halfway vector between the incident directions $\omega_i$ and outgoing directions $\omega_o$. $D_{\mathrm{GGX}}$, $F$, and $G$ are the GGX normal-distribution, Schlick Fresnel, and Smith geometry terms. In this work, the modeled human materials are treated as dielectrics, we fix the normal-incidence Fresnel reflectance to $F_0=0.04$ rather than predicting metallic.

\begin{figure*}[!t]
    \centering
    \includegraphics[width=0.88\textwidth]{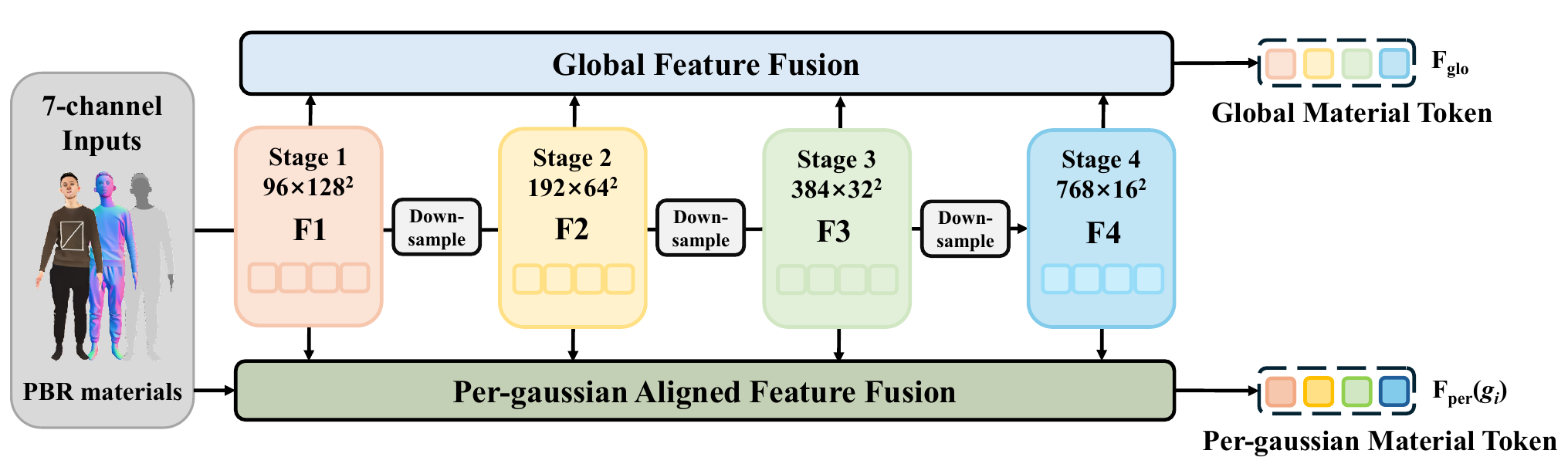}
    \caption{Details of the material encoder. A seven-channel material prior is processed by four ConvNext-style multi-scale encoding blocks, whose features are fused into global material tokens and per-Gaussian material tokens.}
    \label{fig:material-encoder}
\end{figure*}
\subsection{3D Gaussian Splatting}

We adopt the 3D Gaussian Splatting framework~\cite{3DGS} as the 3D representation for avatars. Each Gaussian primitive is parameterized by a position \(\mathbf{p}\), scaling factors \(\mathbf{s}\), rotations \(\mathbf{q}\), opacity \(o\), and spherical harmonic color coefficients \(\mathbf{c}\). Different from vanilla 3DGS, each Gaussian primitive in ZARA is further equipped with albedo \(\mathbf{a}\), roughness \(\mathbf{r}\), and normal \(\mathbf{n}\) parameters to support physically based rendering. Given these properties, the rendering process is given by
\begin{equation}
    \mathbf{I}
    =
    \operatorname{Splatting Rasterizer}
    \left(
        \mathbf{p},
        \mathbf{x},
        \mathbf{s},
        o,
        \mathbf{q}
    \right),
    \label{eq:gaussian-splatting}
\end{equation}
where \(\textbf{I}\) is the rendered image, and $\textbf{x}$ can be \(\mathbf{c}\), \(\mathbf{a}\), \(\mathbf{r}\), and \(\mathbf{n}\). At inference time, we additionally perform physically based rendering to obtain relighting results via deferred shading \cite{Hargreaves2004DeferredShading,DeferredGS}.

\section{Method}

As shown in Fig.~\ref{fig:pipeline-overview}, given a single image, ZARA reconstructs a canonical 3D Gaussian avatar that can be animated and relighted under novel illumination. A pretrained human reconstruction backbone predicts canonical Gaussian geometry, while the proposed material encoder and material transformer estimate per-Gaussian albedo and roughness. Gaussian-aligned feature fusion injects local material evidence from image space to the corresponding Gaussian primitives, and the proposed MatFace Regularization strategy improves facial albedo learning. Given a target motion sequence, the canonical avatar is deformed and rendered to each target pose and illumination. The entire reconstruction process is feed-forward and requires neither person-specific optimization nor test-time fine-tuning. 

\subsection{Geometry Backbone}

We build ZARA on the pretrained LHM~\cite{LHM} backbone. A body encoder and a face encoder extract reference-image tokens $F_b^r$ and $F_h^r$. Canonical 3D query points are sampled from a subdivided SMPL-X template~\cite{SMPLX} and embedded and projected into Gaussian queries. The geometry transformer predicts a latent feature for every query,
\begin{equation}
    T_{geo} = \mathbf{T_G}\bigl(E_q(Q),[F_b^r;F_h^r];\mathbf{e}^r\bigr),
\end{equation}
where $\mathbf{e}^r$ is an image-derived global conditioning vector. We retain the pretrained image encoders and adapt the Geometry Transformer with LoRA~\cite{LoRA}. MLP-based geometry decoder and heads decode $T_{geo}$ into canonical position offsets, scales, rotations, opacities, normal offsets, and SH RGB colors. For positions and normals, we use the canonical SMPL-X parameters as a stable geometric basis, each head of these two attributes predicts a offset from the geometry token $T_{geo}$:
\begin{equation}
\begin{aligned}
\mathbf{p}_{i}^{c}
&= \mathbf{p}_i^{\rm SMPLX}+\Delta\mathbf{p}_i^{\rm head},\\
\mathbf{n}_{i}^{c}
&= \frac{\mathbf{n}_i^{\rm SMPLX}+\Delta\mathbf{n}_i^{\rm head}}
        {\left\|\mathbf{n}_i^{\rm SMPLX}+\Delta\mathbf{n}_i^{\rm head}\right\|_2}.
\end{aligned}
\end{equation}
Together, these design choices retain the large reconstruction model’s geometric and appearance priors while adapting its Gaussian distribution to our training, and also allowing body-oriented normals to capture clothing folds and facial geometry beyond the template.

Static canonical attributes alone cannot represent pose-dependent cloth deformation. Following \cite{DynaAvatar} and \cite{LCA}, we therefore attach a shared non-rigid geometry decoder with separate MLP and heads for position, rotation, scale, and normal corrections. The decoder takes $T_{geo}$ as input, however, is also conditioned on a local window of target SMPL-X poses and on facial expression, jaw, and eye parameters. The resulting corrections are applied in canonical space before linear blend skinning. The geometry decoder reconstructs the canonical Gaussian geometry from the image-conditioned features, while the offset decoder further predicts pose-dependent geometric corrections conditioned on the target pose.

\begin{figure}[!t]
    \centering
    \includegraphics[width=\columnwidth]{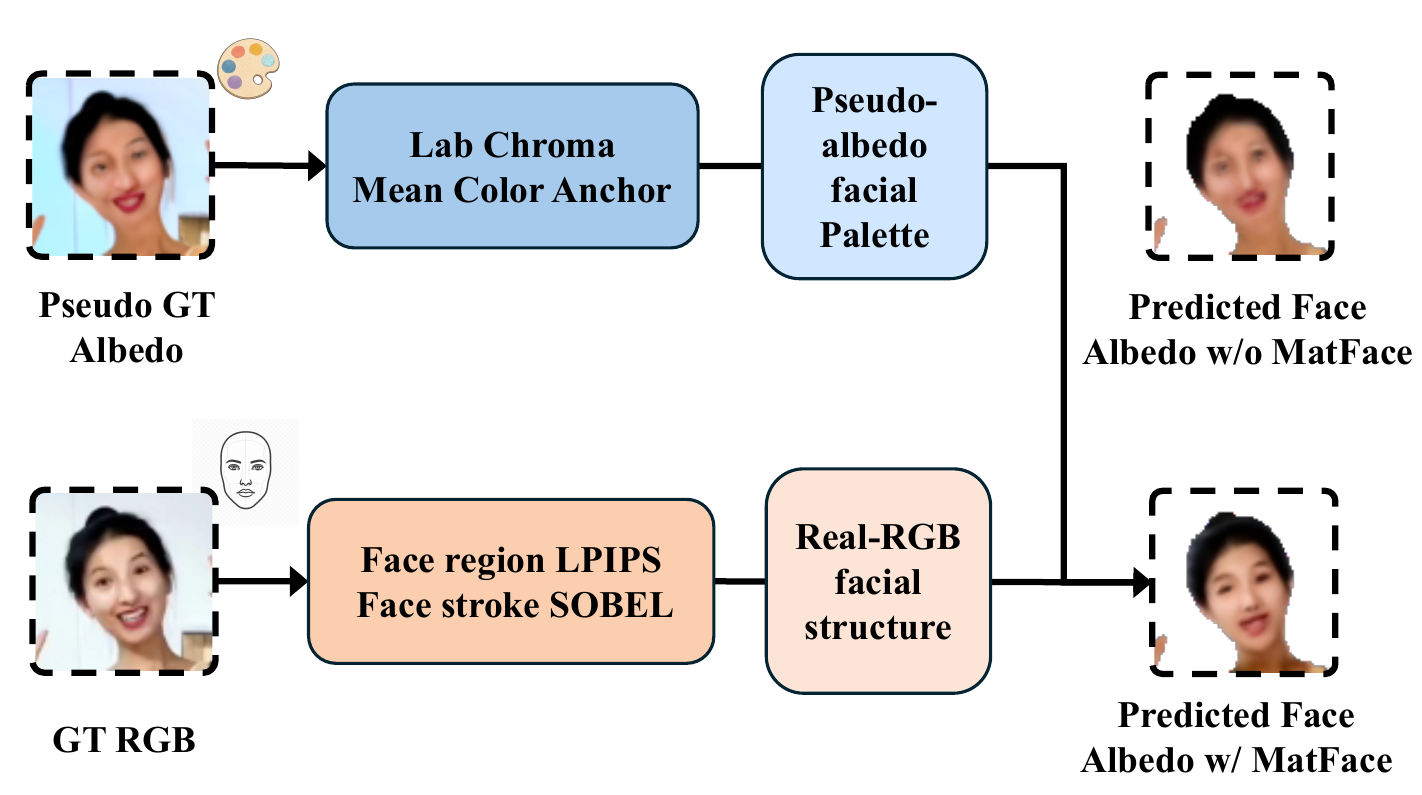}
    \caption{Schematic of MatFace Regularization. The captured RGB face supplies facial structure, while pseudo-label albedo supplies the intrinsic color palette. Their complementary cues supervise the predicted albedo face region during training.}
    \label{fig:matface}
\end{figure}

\subsection{Material Learning, Encoding, and Prediction}
\label{sec:gaussian-material}

\textbf{Material Priors and Supervision.}
A straightforward approach for material learning follows the inverse-rendering paradigm adopted in prior works\cite{TensoIR, GSIR, Relightable3DGaussians, SGIA, DNFAvatar}: the model reconstructs or predicts the material properties of the avatar, produces RGB images via the rendering equation, and performs self-supervision against the ground-truth RGB images. Nevertheless, this scheme requires simultaneous prediction of geometry, materials and lighting, which suffers from the classic ill-posed problem of inverse-rendering. An alternative strategy is to supervise model learning directly using ground-truth materials, which provides richer direct material information for the model. However, existing human-avatar datasets generally lack material annotations. Thanks to advances in video diffusion models (VDMs), VDM-based material generation models are capable of producing high-quality and temporally consistent material maps. We then estimate pseudo-material maps using a frozen VDM-based inverse-rendering model~\cite{DiffusionRenderer}. When the predicted roughness is spatially uniform over the foreground, we replace it with a map constructed by assigning predefined part-specific roughness values to Sapiens2~\cite{Sapiens2} segmentation results from the same RGB image. The resulting reference-view maps provide conditioning, while target-view maps provide supervision. At inference, the same material-prior preparation pipeline operates automatically on the single input RGB image. Alternatively, users can supply edited material maps to control the reconstructed materials.

\textbf{Multi-scale global material encoding.}
As shown in Fig.~\ref{fig:material-encoder}, the masked reference-view albedo, roughness, and normal maps are concatenated into a seven-channel tensor. A multi-scale ConvNeXt encoder~\cite{ConvNeXt} extracts four feature maps, denoted by $\{F_\ell^r\}_{\ell=1}^{4}$. After being aligned, features from all four stages are concatenated and projected into a sequence of global material tokens $\mathbf{F}_{\mathrm{glo}}$. These tokens encode long-range material context, however, they do not establish an explicit correspondence between image regions and individual Gaussian primitives.

\textbf{Gaussian-aligned material feature fusion.}
For each Gaussian $i$, we project its center $\mathbf{p}_i$ into the reference view and bilinearly sample the multi-scale feature maps to obtain center-aligned evidence $\{F_\ell^r(\mathbf{p}_i)\}_{\ell=1}^{4}$. In parallel, following GWFBP \cite{GWFBP}, we apply a splat-aware soft lift to aggregate pseudo material maps over the Gaussian's projected footprint, producing a footprint-aligned observation $\overline{\mathbf{m}}_i^r$. We use these different projection strategies because the center-aligned evidence preserves spatially localized appearance cues, whereas footprint aggregation pools observations over each Gaussian’s projected support and accounts for occlusion and overlapping splats through rendering-contribution weighting. As a result, these complementary features are fused through learned projections into a Gaussian-aligned material feature,
\begin{equation}
    \mathbf{F}_{\mathrm{per}}^i =
    \Phi_{\mathrm{fuse}}\!\left(
    \{F_\ell^r(\mathbf{p}_i)\}_{\ell=1}^{4},
    \overline{\mathbf{m}}_i^r
    \right).
\end{equation}

\textbf{Material Transformer.}
The fused Gaussian-aligned feature is injected into the geometry token through a learned validity-aware gate,
\begin{equation}
    \widetilde{T}_{geo} =
    \operatorname{GatedInject}\!\left(
    T_{geo},\mathbf{F}_{\mathrm{per}}^i;\mathbf{V}
    \right),
\end{equation}
where $\mathbf{V}$ denotes the observation-validity mask. The resulting Gaussian tokens $\widetilde{T}_{geo}$ act as queries in the Multimodal Transformer\cite{Esser2024RectifiedFlow} based block, while the global material tokens provide keys and values. The reference-image global conditioning vector $\mathbf{e}^r$ modulates each block through adaptive layer normalization, yielding the material latent
\begin{equation}
    T_{mat} =
    \mathbf{T_{M}}\!\left(
    \widetilde{T}_{geo},
    \mathbf{F}_{\mathrm{glo}};
    \mathbf{e}^{r}
    \right).
\end{equation}

\textbf{Attribute prediction.}
Albedo and roughness are decoded from $T_{mat}$ by a shared material MLP followed by attribute-specific linear heads,
\begin{equation}
    \mathbf{a}_i =
    \sigma\!\left(h_a\!\left(\phi_m(T_{mat}^i)\right)\right),
    \qquad
    r_i =
    \sigma\!\left(h_r\!\left(\phi_m(T_{mat}^i)\right)\right).
\end{equation}
After initialization, the two heads are optimized independently. During training, the predicted per-Gaussian intrinsic attributes are rendered at the target poses and views for supervision using the corresponding pseudo-material maps.

\subsection{MatFace Regularization}
\label{sec:matface}

Facial region pseudo-label albedo data can exhibit identity and spatial shifts, making direct pixel-wise supervision unreliable. We therefore introduce MatFace Regularization, which decouples facial structure from color: the captured RGB face provides the spatial organization of facial features, while the pseudo-label albedo provides the intrinsic color palette. This training-only supervision design is illustrated in Fig.~\ref{fig:matface}, which is achieved by two carefully designed groups of losses.

We denote $\mathbf{A}_f$, $\mathbf{I}_f$, and $\widetilde{\mathbf{A}}_f$ as the predicted albedo, captured RGB, and pseudo-label albedo, respectively, and all the images are cropped to the facial region. The MatFace Regularization objective is
\begin{equation}
\begin{aligned}
\mathcal{L}_{\rm MatFace}
&=\lambda_{\mathrm{chr}}\mathcal{L}_{\mathrm{chr}}
(\mathbf{A}_f,\mathbf{I}_f,\widetilde{\mathbf{A}}_f) \\
&\quad+\lambda_{\mathrm{col}}\mathcal{L}_{\mathrm{col}}
(\mathbf{A}_f,\widetilde{\mathbf{A}}_f)\\
&\quad+\lambda_{\mathrm{perc}}\mathcal{L}_{\mathrm{perc}}
(\mathbf{A}_f,\mathbf{I}_f) \\
&\quad+\lambda_{\mathrm{edge}}\mathcal{L}_{\mathrm{edge}}
(\mathbf{A}_f,\mathbf{I}_f) .
\end{aligned}
\end{equation}
Specifically, let \(\mathbf{A}_f^{ab}\), \(\mathbf{I}_f^{ab}\), and \(\widetilde{\mathbf{A}}_f^{ab}\) denote the CIELAB \(a,b\) chromaticity maps obtained from the predicted-albedo, input-RGB, and pseudo-label albedo face crops, respectively. The chromaticity loss is
\begin{equation}
\mathcal{L}_{\mathrm{chr}}
=\mathcal{L}_{1}\!\left(
\mathbf{A}_f^{ab},\mathbf{T}_f^{ab}\right).
\label{eq:matface-chroma}
\end{equation}
Here, $\mathcal{L}_{1}(\cdot,\cdot)$ denotes the mean L1 loss, and $\mathbf{T}_f^{ab}$ is the hybrid chromaticity target
\begin{equation}
\begin{aligned}
\mathbf{T}_f^{ab}(p)={}&\mu_M(\widetilde{\mathbf{A}}_f^{ab})
+\sigma_M(\widetilde{\mathbf{A}}_f^{ab})\odot{}\\
&\operatorname{clip}_3\!\left(
\frac{\mathbf{I}_f^{ab}(p)-\mu_M(\mathbf{I}_f^{ab})}
{\sigma_M(\mathbf{I}_f^{ab})}\right).
\end{aligned}
\label{eq:matface-chroma-target}
\end{equation}
where $p$ indexes pixels in the face crop, while $\mu_M$ and $\sigma_M$ denote the channel-wise mask-weighted mean and standard deviation, respectively, and $\operatorname{clip}_3$ clips each value to $[-3,3]$. The input RGB provides spatially aligned relative chromatic variations but contains illumination-dependent color, whereas pseudo-albedo provides a more intrinsic chromatic palette but may suffer from facial misalignment. This construction therefore normalizes the RGB chroma and rescales it using the pseudo-label albedo statistics, combining the reliable spatial pattern of the former with the intrinsic palette of the latter. Clipping further suppresses outlier values.

The mean-color loss is
\begin{equation}
\mathcal{L}_{\mathrm{col}}
=\frac{1}{3}\left\|
\mu_M(\mathbf{A}_f)-
\mu_M(\widetilde{\mathbf{A}}_f)\right\|_1.
\label{eq:matface-color}
\end{equation}
This loss pulls the overall color of the predicted albedo toward the pseudo‑label.

To transfer the stable facial edge structure of the input RGB to the predicted albedo, we employ $\mathcal{L}_{\mathrm{perc}}$, which computes LPIPS~\cite{LPIPS} between independently normalized predicted-albedo and RGB face crops. Furthermore, we define a spatially tolerant edge L1 loss as
\begin{equation}
\begin{aligned}
\mathcal{L}_{\mathrm{edge}}
=\mathcal{L}_{1}^{\mathrm{tol}}\big(&
\operatorname{Sobel}(\mathcal{H}(\mathbf{A}_f)),\\
&\operatorname{Sobel}(\mathcal{H}(\mathbf{I}_f))\big).
\end{aligned}
\label{eq:matface-edge}
\end{equation}
Here, $\mathcal{H}(\cdot)$ extracts high-pass log luminance by converting RGB to log luminance and subtracting its $11\times11$ local average. This suppresses low-frequency shading while retaining facial structure for albedo supervision. $\operatorname{Sobel}(\cdot)$ denotes the Sobel operation. $\mathcal{L}_{1}^{\mathrm{tol}}$ denotes a spatially tolerant L1 distance that compares each Sobel response with the strongest response in a $5\times5$ neighborhood of the other map. Unlike standard L1, which compares responses at the same pixel, this formulation tolerates small spatial displacements while retaining penalties for missing or spurious edges, thereby preventing minor misalignments from blurring facial textures. With a zero neighborhood radius, $\mathcal{L}_{1}^{\mathrm{tol}}$ reduces to standard L1. 

\subsection{Training Objectives}

Our overall objective is
\begin{eqnarray}
    \mathcal{L} &=& \lambda_a\mathcal{L}_{a}
    +\lambda_r\mathcal{L}_{r}
    +\lambda_n\mathcal{L}_{n}
    +\lambda_m\mathcal{L}_{\rm mask} \nonumber\\
    &&+\lambda_f\mathcal{L}_{\rm MatFace}
    +\lambda_{\rm reg}\mathcal{L}_{\rm reg}
    +\lambda_{\rm rgb}\mathcal{L}_{\rm rgb}.
\end{eqnarray}
The albedo loss \(\mathcal L_a\) combines L1, MSG \cite{MVInverse}, and LPIPS, while the roughness \(\mathcal L_r\) and normal \(\mathcal L_n\) losses use L1 and MSG. The sub losses of these loss are summed without additional weighting. $\mathcal{L}_{a}$ is performed outside the face region, whereas facial albedo is supervised by $\mathcal{L}_{\rm MatFace}$. We define $\mathcal{L}_{\rm mask}=\mathcal{L}_{1}(\widehat{\mathbf{M}},\mathbf{M})$ as the mean L1 loss between the rendered accumulated opacity $\widehat{\mathbf{M}}$ and the binary foreground mask $\mathbf{M}$. Regularization $\mathcal{L}_{\rm reg}$ includes ASAP/ACAP from LHM~\cite{LHM} and magnitude penalties on the applied pose-dependent geometry decoder corrections. We retain the RGB loss to facilitate the reconstruction of geometry and fine-grained color details. This loss combines foreground L1 and LPIPS supervision between the RGB output from the original LHM\cite{LHM} (geometry backbone) and the captured RGB image. 

We set \(\lambda_a=\lambda_r=\lambda_n=\lambda_m=\lambda_f=\lambda_{\mathrm{reg}}=1\), and \(\lambda_{\mathrm{rgb}}=0.2\), with individual regularization weights incorporated into  \(\mathcal{L}_{\mathrm{MatFace}}\) and \(\mathcal{L}_{\mathrm{reg}}\). Within \(\mathcal{L}_{\mathrm{MatFace}}\), the $\lambda_{\mathrm{chr}}$, $\lambda_{\mathrm{col}}$, $\lambda_{\mathrm{perc}}$, and $\lambda_{\mathrm{edge}}$ are set to 0.2, 1, 1, and 1, respectively. And for \(\mathcal{L}_{\mathrm{reg}}\), the ASAP and ACAP terms are weighted by 10 and 20, respectively, while the magnitude penalties on pose-dependent position, rotation, scale, and normal corrections are weighted by 1, 0.1, 0.5, and 0.1, respectively.

\begin{table*}[!t]
\caption{Quantitative comparison of material prediction on Human4DiT-3D and RANA. Person-specific methods ($\dagger$) use full subject-specific videos and iterative optimization while zero-shot methods uses a single image without test-time optimization. For the zero-shot baselines, we report albedo metrics and normal predictions are unavailable. The best result for each metric is bolded separately within each group.}
\label{tab:material-comparison}
\centering
\footnotesize
\setlength{\tabcolsep}{5.0pt}
\renewcommand{\arraystretch}{1.08}
\begin{tabular}{@{}l*{8}{c}@{}}
\hline
& \multicolumn{4}{c}{Human4DiT-3D \cite{Human4DiT}} & \multicolumn{4}{c}{RANA \cite{RANA}} \\
\cline{2-5}\cline{6-9}
Method
& \multicolumn{3}{c}{Albedo}
& Normal
& \multicolumn{3}{c}{Albedo}
& Normal \\
\cline{2-4}\cline{5-5}\cline{6-8}\cline{9-9}
& PSNR $\uparrow$ & SSIM $\uparrow$ & LPIPS $\downarrow$
& MAE($^\circ$) $\downarrow$
& PSNR $\uparrow$ & SSIM $\uparrow$ & LPIPS $\downarrow$
& MAE($^\circ$) $\downarrow$ \\
\hline
\multicolumn{9}{@{}l}{\textbf{Available data: full subject-specific video}} \\
IntrinsicAvatar$^\dagger$ \cite{IntrinsicAvatar} & \textbf{21.853} & \textbf{0.8362} & 0.0845 & \textbf{9.521} & \textbf{22.696} & \textbf{0.8730} & \textbf{0.0541} & 10.781 \\
DNF-Avatar$^\dagger$ \cite{DNFAvatar} & 15.439 & 0.7003 & \textbf{0.0738} & 17.785 & 22.243 & 0.8647 & 0.0567 & \textbf{10.508} \\
\hline
\multicolumn{9}{@{}l}{\textbf{Available data: a single image}}\\
LHM \cite{LHM} + UniRelight \cite{UniRelight} & 18.633 & 0.7600 & 0.0733 & -- & 19.54 & 0.8447 & 0.1296 & --\\
IDOL \cite{IDOL} + UniRelight \cite{UniRelight} & 17.639 & 0.7439 & 0.0790 & -- & 18.82 & 0.8405 & 0.1291 & --\\
ZARA (Ours) & \textbf{22.575} & \textbf{0.8974} & \textbf{0.0320} & 13.703 & \textbf{20.931} & \textbf{0.8494} & \textbf{0.0698} & 14.965\\
\hline
\end{tabular}
\end{table*}

\begin{table*}[!t]
\caption{Quantitative comparison of relighting RGB on Human4DiT-3D and RANA. Person-specific methods ($\dagger$) use full subject-specific videos and iterative optimization, whereas zero-shot methods use a single image. The best result for each metric is bolded separately within each group.}
\label{tab:relit-comparison}
\centering
\footnotesize
\setlength{\tabcolsep}{4.5pt}
\renewcommand{\arraystretch}{1.08}
\begin{tabular}{@{}lccccccccc@{}}
\hline
& \multicolumn{4}{c}{Human4DiT-3D \cite{Human4DiT}} & \multicolumn{4}{c}{RANA \cite{RANA}} & \\
\cline{2-5}\cline{6-9}
Method
& PSNR $\uparrow$ & SSIM $\uparrow$ & LPIPS $\downarrow$ & FC $\downarrow$
& PSNR $\uparrow$ & SSIM $\uparrow$ & LPIPS $\downarrow$ & FC $\downarrow$
& Runtime $\downarrow$ \\
\hline
\multicolumn{10}{@{}l}{\textbf{Available data: full subject-specific video}} \\
IntrinsicAvatar$^\dagger$ \cite{IntrinsicAvatar} & \textbf{20.373} & \textbf{0.8276} & 0.0546 & \textbf{0.0461} & 18.332 & 0.8693 & \textbf{0.0367} & 0.0368 & \textbf{2.3~h} \\
DNF-Avatar$^\dagger$ \cite{DNFAvatar} & 17.372 & 0.7441 & \textbf{0.0534} & 0.0474 & \textbf{18.935} & \textbf{0.8715} & 0.0510 & \textbf{0.0363} & 6.6~h \\
\hline
\multicolumn{10}{@{}l}{\textbf{Available data: a single image}} \\
LHM \cite{LHM} + UniRelight \cite{UniRelight} & 14.949 & 0.6330 & 0.1353 & 0.0892 & 15.413 & 0.6636 & 0.2040 & 0.0974 & 101.70~s \\
IDOL \cite{IDOL} + UniRelight \cite{UniRelight} & 13.145 & 0.6019 & 0.1510 & 0.1206 & 14.358 & 0.6957 & 0.1814 & 0.1104 & 97.23~s \\
ZARA (Ours) & \textbf{20.667} & \textbf{0.8605} & \textbf{0.0427} & \textbf{0.0425} & \textbf{18.550} & \textbf{0.8542} & \textbf{0.0650} & \textbf{0.0502} & \textbf{21.05~s} \\
\hline
\end{tabular}
\end{table*}

\section{Experiments}

\subsection{Datasets collection}
\label{sec:material-data}

For training, we use Human4DiT-3D \cite{Human4DiT}, DNA-Rendering \cite{DNARendering} and in-the-wild human videos from YouTube and BiliBili which is collected by Human4DiT \cite{Human4DiT}. Total training data comprise approximately 1.8 million frames. Human4DiT‑3D provides scanned assets, yet only meshes and textures captured under uniform illumination are available. We feed each asset into the PBR pipeline and render it with a specified environment map as the input and reference to our model. For quantitative evaluation, we render the original uniform‑lighting textures of Human4DiT-3D as the albedo ground truth and the geometric normals as the normal ground truth. We also perform test experiments on RANA \cite {RANA} datasets, which is a synthetic dataset that includes ground-truth albedo, normal, and relighting results at novel pose for evaluation. For Human4DiT-3D and DNA-Rendering, we split the dataset so that training and testing subjects do not overlap, obtaining 50 and 10 test subjects, respectively. RANA dataset are randomly sampled 10 subjects for testing. We also collect additional 50 in-the-wild human videos from BiliBili for qualitative testing. For each sample, we generate pseudo-material maps using the pipeline described in Sec.~\ref{sec:gaussian-material} during training. The resulting maps provide material priors for reference observations and target supervision during training. 

\begin{figure*}[!tp]
    \centering
    \includegraphics[width=\textwidth]{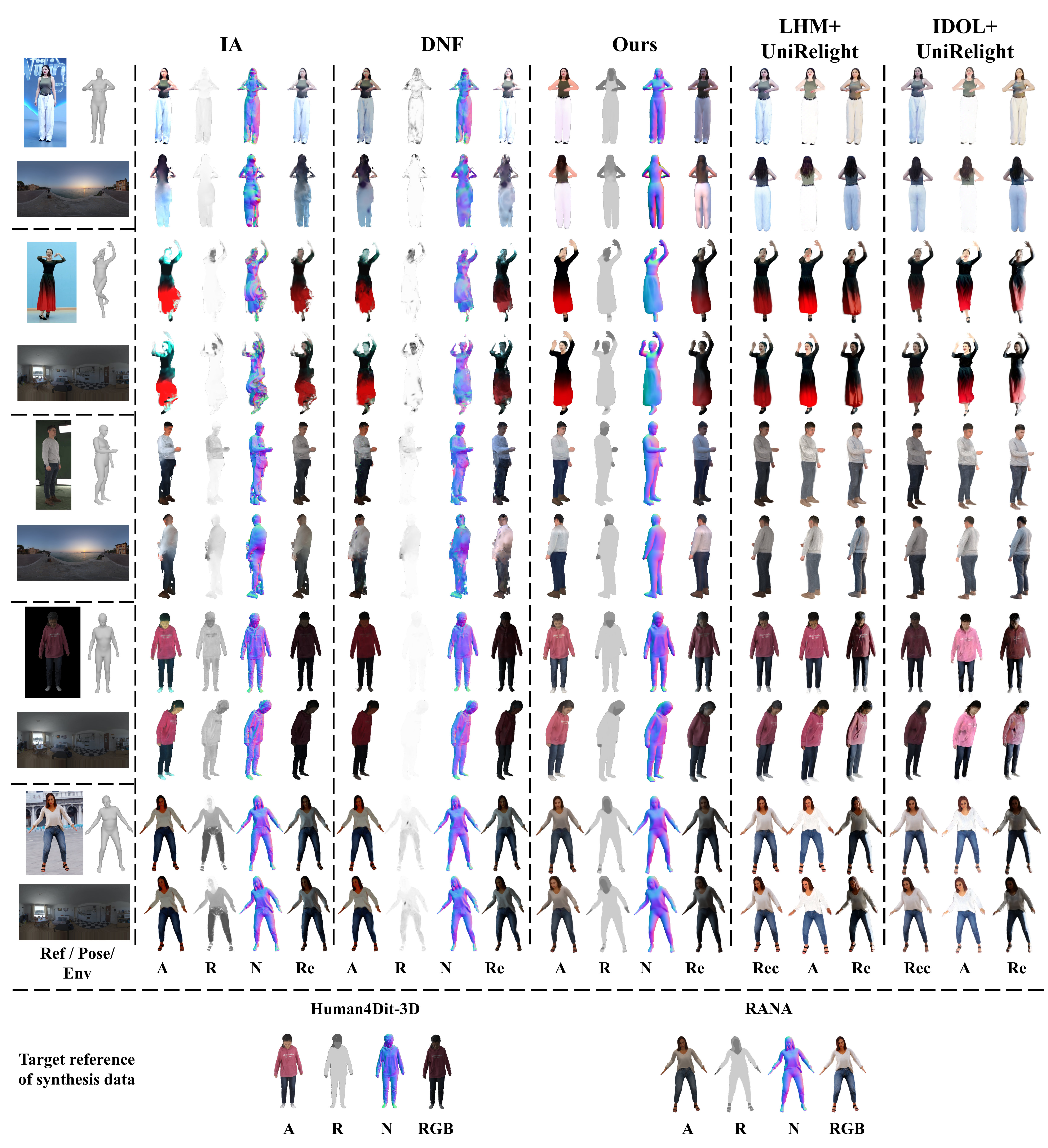}
    \caption{Qualitative comparison between ZARA, zero-shot baseline methods and person-specific methods with in-the-wild video (the first two samples), studio-captured video (third sample), and synthesis data (the last two samples). For each sample, the first row shows the input view and the second row shows the unseen view. Ref, Pose, Env, A, R, N, Re, and Rec are the reference image, target pose, environment map for relighting, albedo, roughness, normal, relighting rgb, and reconstructed rgb, respectively. For the two-stage zero-shot baselines, we show reconstructed RGB, estimated albedo, and relighting results. For ZARA and person-specific approaches, we visualize the material maps and relighting outputs. Below the dashed line at the bottom are references ground truth from the synthetic dataset, including material maps and RGB images.}
    \label{fig:qualitative-comparison}
\end{figure*}

\subsection{Implementation Details}

Training is conducted on 8 NVIDIA H200 GPUs and requires approximately 7 days. Both the source images and supervision targets are processed at a resolution of $512\times512$. We use the Adam optimizer with $\beta_1=0.9$, $\beta_2=0.999$, zero weight decay, and a maximum gradient norm of 1.0. Module-specific learning rates are set to $1\times10^{-4}$ for the material MLP and heads, the material-encoder, and the pose-dependent geometry decoder; $2\times10^{-4}$ for the LoRA parameters and Material Transformer; and $4\times10^{-4}$ for the remaining trainable modules. The Geometry Transformer is adapted using rank-32 LoRA~\cite{LoRA} with an alpha of 64 and a dropout rate of 0.1. 

\subsection{Evaluation Metrics}

Quantitative material evaluation is conducted on held-out Human4DiT-3D and RANA dataset using their reference albedo and normal maps. We test albedo on Peak Signal-to-Noise Ratio (PSNR), Structural Similarity (SSIM)~\cite{SSIM}, and Learned Perceptual Image Patch Similarity LPIPS~\cite{LPIPS}. For normals, we test the Mean Angular Error (MAE). We additionally evaluate relit RGB images using PSNR, SSIM, LPIPS, and Face Consistency (FC) \cite{LHM}, where the environment map is held fixed across tests of each sample. All image-space metrics are computed within the valid foreground region. DNA-Rendering and in-the-wild videos are used for qualitative evaluation of real-world material generalization and relighting.

%\FloatBarrier
\subsection{Comparison with Baseline Methods}
\label{sec:comparison}
\subsubsection{Baselines Selection}
As there is, to the best of our knowledge, no existing open-source method supports feed-forward reconstruction of an animatable and relightable avatar from a single image in a zero-shot manner, we further construct two feed-forward avatar-generation baselines by combining LHM~\cite{LHM} and IDOL~\cite{IDOL} with the State of the Art (SOTA) video relighting model UniRelight~\cite{UniRelight}. These two-stage baselines first generate and animate an avatar from the reference image and then estimate albedo and relighting results. The same video relighting method and inference settings are used for both avatar generators so that their differences are not confounded by different relighting networks. 

Following the comparison strategy in PERSONA \cite{PERSONA}, we also test two SOTA person-specific relightable avatar reconstruction methods, IntrinsicAvatar \cite{IntrinsicAvatar} and DNF-Avatar~\cite{DNFAvatar} as reference. A direct comparison between zero-shot methods and person-specific methods is unfair, as the former take a single image as input and require no test-time optimization, while the latter use full subject-specific training sequences and perform iterative optimization for each identity. We retain this distinction rather than treating the person-specific methods as baselines operating on a single image. On the one hand, this vanilla setting allows the person-specific methods to fully exploit their capabilities. On the other hand, it enables us to assess whether our zero-shot model can rapidly achieve performance comparable to that of methods requiring several hours of per-subject training. We both test the materials reconstruction and relighting results for these two methods. 

\begin{figure*}[!tb]
    \centering
    \includegraphics[width=0.95\textwidth]{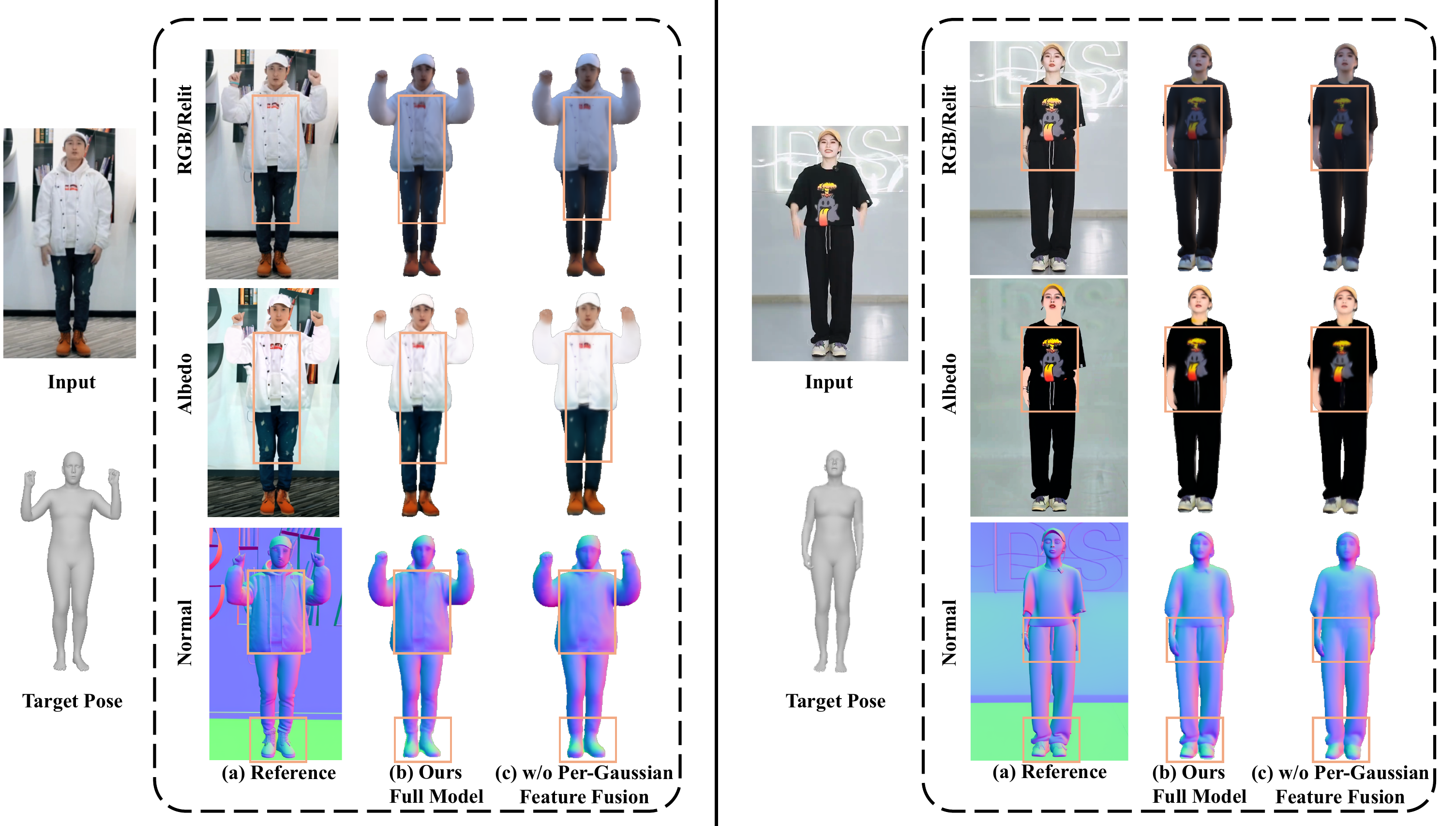}
    \caption{Qualitative ablation of Gaussian-aligned material fusion on two real-world samples. For each sample, the reference image and target pose are followed by RGB/Relit, albedo, and normal results from the full model and the ablated model. The \textcolor{orange}{orange boxes} indicate localized garment regions used for comparison.}
    \label{fig:per-gaussian-ablation}
\end{figure*}

\subsubsection{Quantitative Comparison}

Tables~\ref{tab:material-comparison} and~\ref{tab:relit-comparison} report the material and relighting comparison results, respectively between ZARA and baseline methods and person-specific references. All experiments are conducted on a single H200 GPU. Runtime includes material-prior estimation, input encoding, avatar reconstruction, and rendering/relighting for zero-shot methods. And for person-specific methods, it denotes per-subject optimization time. Compared with the zero-shot baselines, ZARA achieves better performance across all reported metrics while also providing higher inference efficiency. Compared with person-specific reconstruction methods, ZARA obtains comparable material and relighting results on RANA dataset and even surpasses them in PSNR, SSIM, and LPIPS on Human4DiT-3D dataset. Notably, ZARA reaches this level of performance within only tens of seconds per sample, whereas person-specific methods require several hours of iterative optimization for each subject.

\begin{figure*}[!t]
    \centering
    \includegraphics[width=0.86\textwidth]{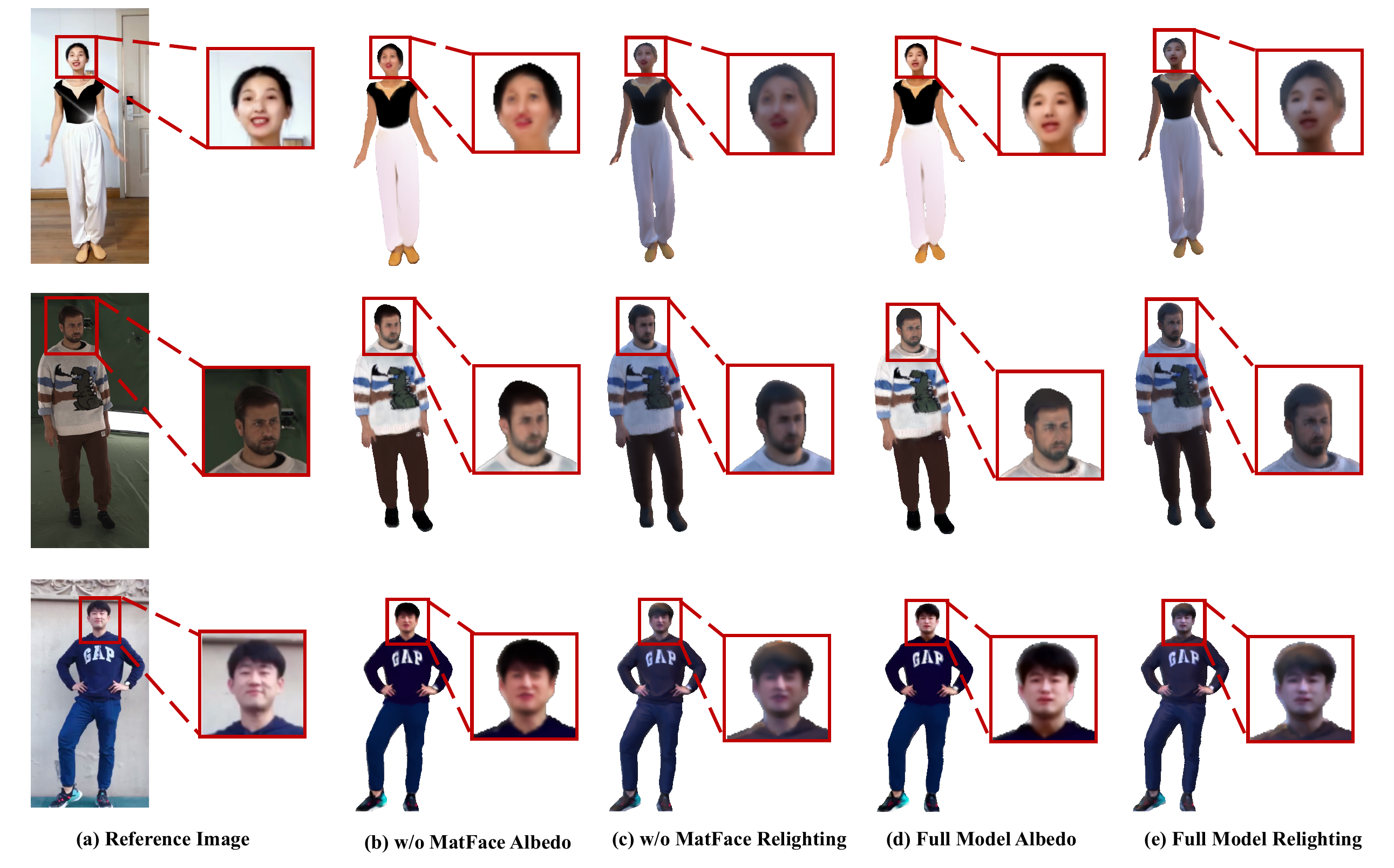}
    \caption{Qualitative ablation of MatFace Regularization. From left to right, we show the reference image, albedo and relighting results without MatFace, and the corresponding full-model results. \textcolor{red}{Red boxes} indicate the facial regions and their enlarged views.}
    \label{fig:matface-ablation}
\end{figure*}

\begin{figure*}[!t]
    \centering
    \includegraphics[width=0.9\textwidth]{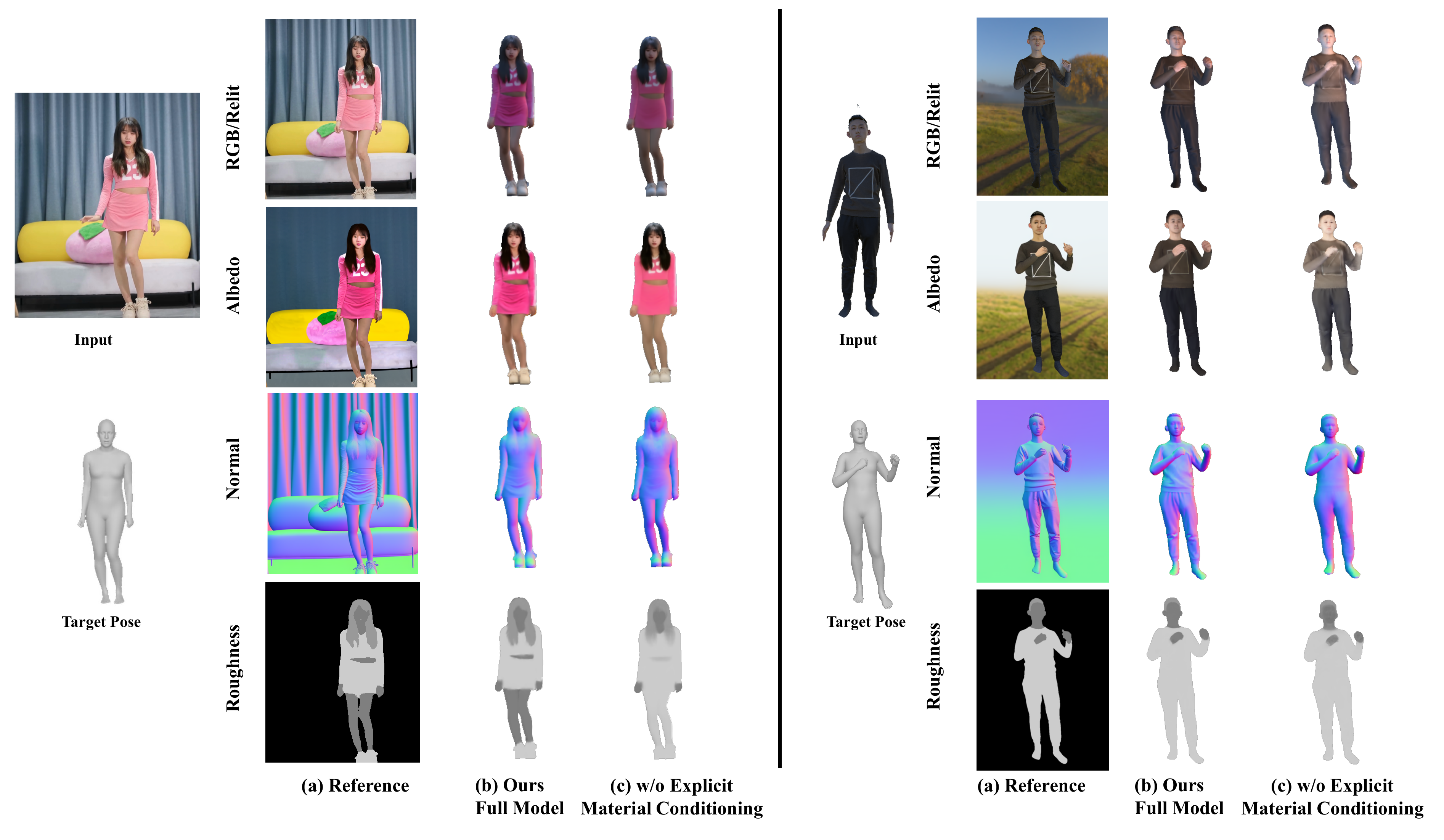}
    \caption{Qualitative ablation of explicit material conditioning. For each sample, the reference image and target pose are followed by intrinsic-material and relighting results from the full model and the RGB-only variant without the explicit material-conditioning pathway.}
    \label{fig:material-transformer-ablation}
\end{figure*}

\subsubsection{Qualitative Comparison}
Fig.~\ref{fig:qualitative-comparison}. presents qualitative comparisons between ZARA and the baseline methods and person-specific references in terms of material estimation and relighting appearance. As the two zero-shot baselines follow two-stage pipelines and the ability of jointly estimating albedo and relighting results by UniRelight \cite{UniRelight}, we therefore show their intermediate reconstructions, albedo results and final relighting results under the specified environment maps. Although both baselines produce plausible reconstructions and albedo results, their relighting outputs exhibit noticeable facial identity drift, overexposure, and view-inconsistent rendering. This behavior arises because the reconstructed 3D avatars retain appearance coupled with the input illumination, whereas the subsequent relight stage performs generative relighting only in image space and lacks explicit identity constraints and material correspondences. Furthermore, we observe inconsistencies in fine details between UniRelight’s jointly generated albedo and relighting results. Although its joint denoising formulation enables cross-modal interaction, it does not explicitly enforce structural consistency between the two outputs. 

On the other hand, although the two person-specific methods yield favorable results on synthetic data, they perform poorly on real-world data in terms of both material estimation and relighting. This discrepancy is mainly because the synthetic setting provides complete single-view 360-degree video sequences rather than observing only a partial view of the body. Real-world sequences, however, may contain large unobserved regions. What's more, although the body and garment details are well reconstructed with synthesis data, both methods also estimate physically implausible roughness values, incorrect brightness, and overly wrinkled normals. These failures cause by the inherent ill-posedness of inverse rendering under RGB self-supervision. ZARA instead learns both human geometry and material attributes from large-scale diverse data, enabling it to predict plausible and consistent material attributes as while as relight results of 3D avatar in both the reference and unseen views from a single image, thereby avoiding the above failure cases of the two-stage hybrid baselines and person-specific reconstruction methods. Besides, we also present a video qualitative comparison results of materials and relighting in the supplemental material.

\subsection{Ablation Studies}
\label{sec:ablation}

\begin{table*}[!t]
\caption{Quantitative ablation on Human4DiT-3D and RANA. Each metric is averaged over the two datasets and the best metrics are bolded.} 
\label{tab:ablation}
\centering
\small
\setlength{\tabcolsep}{3pt}
\begin{tabular}{@{}lcccccccc@{}}
\hline
& \multicolumn{3}{c}{Albedo} & \multicolumn{1}{c}{Normal} & \multicolumn{4}{c}{Relighting RGB} \\
\cline{2-4}\cline{5-5}\cline{6-9}
Variant & PSNR $\uparrow$ & SSIM $\uparrow$ & LPIPS $\downarrow$ & MAE($^\circ$) $\downarrow$ & PSNR $\uparrow$ & SSIM $\uparrow$ & LPIPS $\downarrow$ & FC $\downarrow$ \\
\hline
w/o Gaussian-aligned Proj. & 20.766 & 0.8613 & 0.0665 & 18.681 & 18.232 & 0.8236 & 0.0781 & 0.0539 \\
w/o MatFace Regularization & 21.430 & 0.8640 & 0.0530 & 14.926 & 19.506 & 0.8467 & 0.0578 & 0.0826 \\
w/o Explicit Material Cond. & 18.316 & 0.8256 & 0.0948 & 21.508 & 16.475 & 0.7929 & 0.1010 & 0.0572 \\
Full model & \textbf{21.753} & \textbf{0.8734} & \textbf{0.0509} & \textbf{14.334} & \textbf{19.609} & \textbf{0.8574} & \textbf{0.0539} & \textbf{0.0464} \\
\hline
\end{tabular}
\end{table*}

\subsubsection{Gaussian-Aligned Material Feature Fusion}
We evaluate the influence of Gaussian-Aligned Material Feature Fusion in the following, which is shown in Fig.~\ref{fig:per-gaussian-ablation}. Although the global material tokens from the material encoder provide the Material Transformer with informative material cues, these tokens alone tend to oversmooth regions with variations in color and geometry. Gaussian-Aligned Material Feature Fusion instead equips each Gaussian query with material features from its projected location and local neighborhood, thereby preserving garment textures and fold details from the reference image. The quantitative results in Table~\ref{tab:ablation} further confirm that the Gaussian-Aligned Material Feature Fusion mechanism improves the quality of both material estimation and relighting.

\subsubsection{MatFace Regularization}
To analyze MatFace Regularization, we replace it with the pseudo-label albedo supervision at same frame used for non-face regions, including L1, MSG, and LPIPS losses, while keeping the remaining components and training settings unchanged. As shown in Fig.~\ref{fig:matface-ablation}, removing MatFace Regularization produces blurred and distorted facial albedo, including degraded facial structures, implausible intrinsic colors, and visible identity drift, while these errors also propagate to the relighting results. In contrast, the full model preserves facial structure, intrinsic color, and subject identity more faithfully in the predicted albedo and relighting. The localized comparisons further reinforce the above conclusion. These effects are consistent across the samples from different datasets. Table~\ref{tab:ablation} shows that, removing MatFace Regularization also degrades FC on the synthetic benchmark. Without MatFace Regularization, unreliable facial supervision from pseudo-label adversely affects the learned facial representation and consequently reduces identity consistency.

\begin{figure}[!b]
    \centering
    \includegraphics[width=\columnwidth]{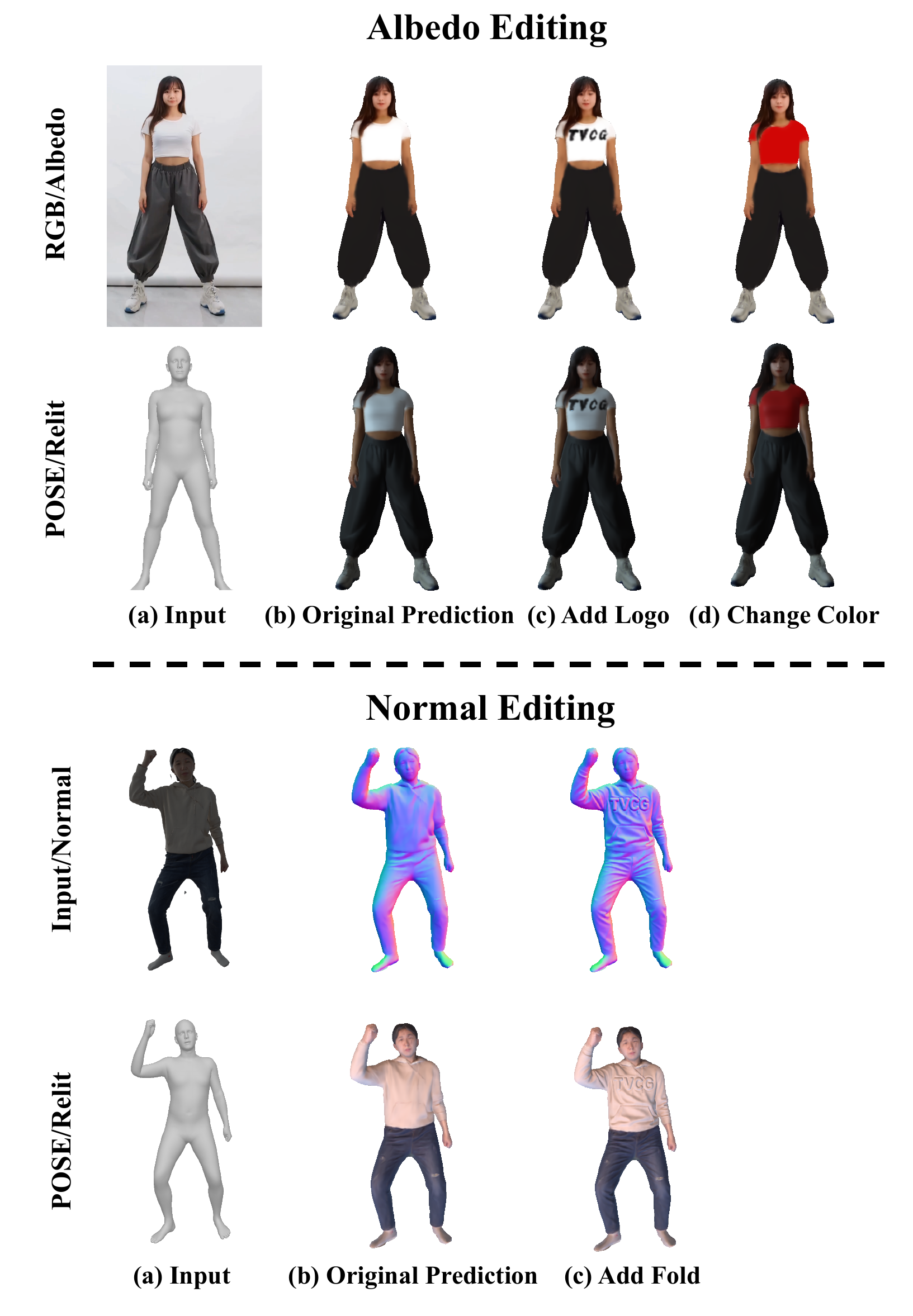}
    \caption{Downstream material editing applications. Top: Albedo editing adds a logo and changes garment colors, with the corresponding relighting results shown below. Bottom: Normal editing changes the distribution of garment folds and the resulting relighting appearance.}
    \label{fig:material-editing}
\end{figure}

\subsubsection{Explicit Material Conditioning}
We investigate the effectiveness of Explicit Material Conditioning. For this module-level ablation, we remove all conditioning pathways that input material cues into the network and investigate whether satisfactory results can still be achieved using only a single RGB image as input while retaining material supervision during training. As shown in Fig~\ref{fig:material-transformer-ablation}, the ablated model without Explicit Material Conditioning reconstructs overly smooth material maps and fails to recover fine-grained texture details. Removing the material-conditioning pathways requires the model to infer intrinsic material properties directly from RGB appearance. In contrast, by injecting both global and local material features, the full model preserves fine-grained details in its predictions. The quantitative results in Table~\ref{tab:ablation} also show performance degradation in the ablated model, further supporting the above conclusion.

%\FloatBarrier
\subsection{Downstream Applications}
ZARA reconstructs a 3D Gaussian avatar together with its intrinsic attributes, providing users with an editable representation for modifying the avatar's appearance. As shown in Fig.~\ref{fig:material-editing}, editing the albedo changes clothing colors and allows logos to be added. Editing the normal map further changes the apparent distribution of garment folds.

\begin{figure}[!t]
    \centering
    \includegraphics[width=\columnwidth]{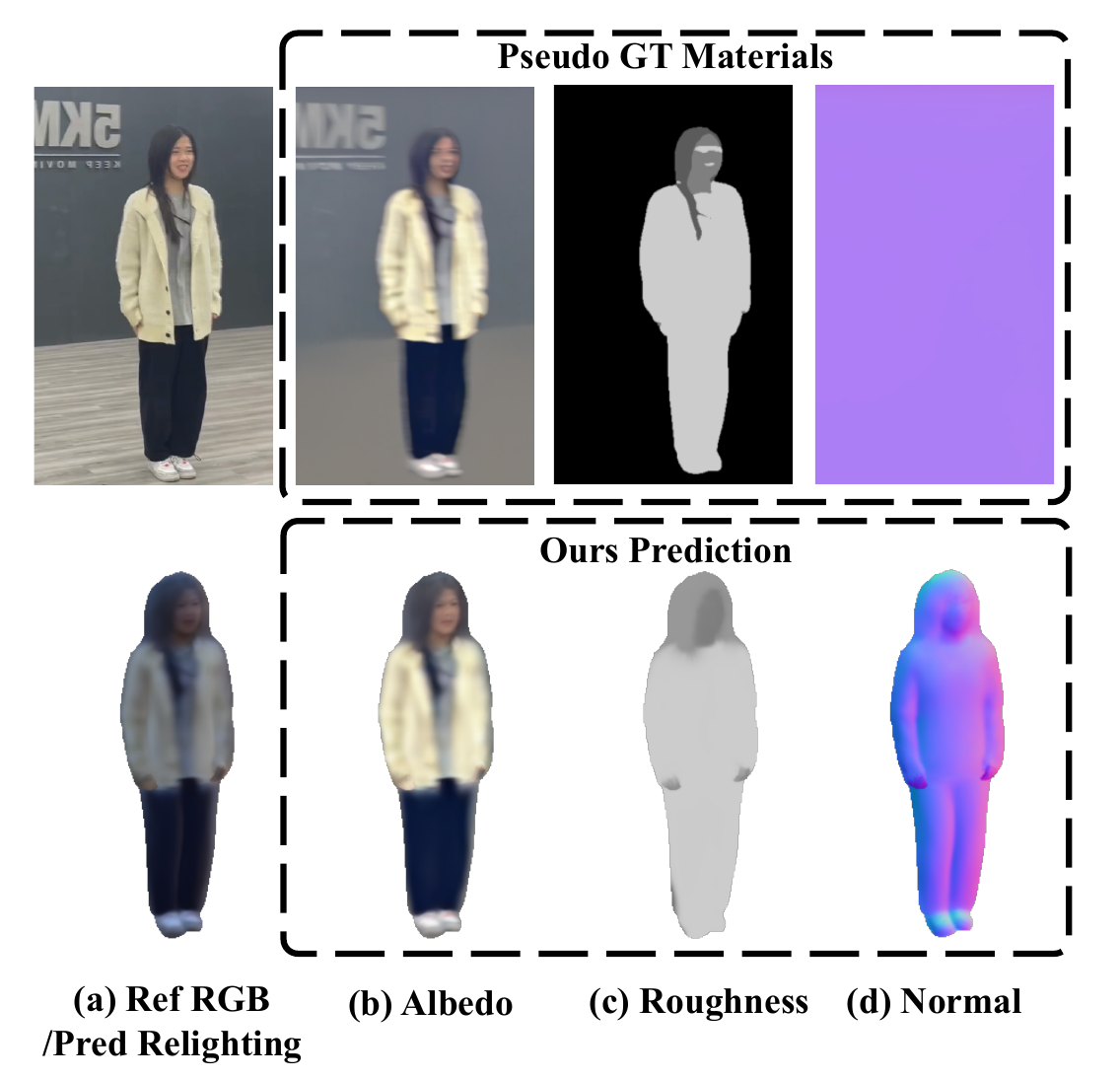}
    \caption{Failure-case under corrupted pseudo-material input. The upper block shows the reference RGB image and erroneous pseudo-albedo, pseudo-roughness, and pseudo-normal maps, while the lower block presents ZARA's relighting appearance and predicted intrinsic materials.}
    \label{fig:pseudo-material-failure-case}
\end{figure}

%\FloatBarrier
\section{Discussion and Conclusion}
\subsection{Limitations}
Although ZARA enables remarkable reconstruction of animatable and relightable avatars from a single image, it still has some limitations. First, as shown in Fig.~\ref{fig:pseudo-material-failure-case}, the reconstruction quality depends on the quality of the pseudo material labels. Errors in the estimated albedo, roughness, or normals can propagate to the predicted Gaussian attributes and become more apparent under novel illumination. Improving robustness to imperfect material inputs is therefore an important direction for future work. Second, our current PBR formulation does not model subsurface scattering, which is important for the soft and translucent appearance of human skin. It is interesting to incorporate an efficient skin-specific scattering model could further improve the realism of faces and other exposed skin regions.

\subsection{Conclusion}
In this paper, we introduce ZARA, a zero-shot feed-forward framework for reconstructing animatable and relightable 3D Gaussian avatars from a single image. By coupling a pretrained Geometry Transformer with a Material Transformer, ZARA predicts explicit per-Gaussian intrinsic attributes for pose-driven animation and deferred PBR. During training, we adopt pseudo-material labels as conditioning inputs and supervision targets. Gaussian-Aligned materials feature fusion preserves local material details, while MatFace Regularization improves facial albedo fidelity under pseudo-label material supervision. Extensive experiments demonstrate that ZARA outperforms zero-shot baselines and achieves results comparable to person-specific methods, while achieving reconstruction time in tens of seconds. The explicit material representation further supports direct editing of avatar appearance.

\bibliographystyle{IEEEtran}
\bibliography{references}

\vfill

\end{document}